\pdfoutput=1

\documentclass[11pt]{article}

\usepackage[]{acl}

\usepackage{times}
\usepackage{float}
\usepackage{amsmath}
\usepackage{amssymb}
\usepackage{bbm}
\usepackage{latexsym}
\usepackage{graphicx}
\usepackage{titlesec}
\usepackage{enumitem}
\usepackage{listings}
\usepackage{comment}
\usepackage{caption}

\usepackage[T1]{fontenc}

\usepackage[utf8]{inputenc}

\usepackage{makecell}
\usepackage{booktabs}
\usepackage{ragged2e}
\usepackage{tabularx}
\usepackage{subcaption}
\usepackage{microtype}

\lstdefinestyle{mystyle}{
    basicstyle=\ttfamily\small,
    numberstyle=\ttfamily\tiny\color{gray},
    breakatwhitespace=true,
    breaklines=true,
    breakautoindent=true,
    breakindent=0pt,
    captionpos=b,
    keepspaces=true,
    numbers=left,
    numbersep=5pt,
    showspaces=false,
    showstringspaces=false,
    showtabs=false,
    tabsize=4,
    frame=single
}

\title{\Large Fill in the Blank: Exploring and Enhancing LLM \\ Capabilities for Backward Reasoning in \\ Math Word Problems }
\author{
    Aniruddha Deb$^{1*}$ \quad
    Neeva Oza$^{1}$\thanks{\hspace{0.3em} Equal Contribution, $^\dagger$Work done while at IIT Delhi. \\Corresponding Authors: \\
    \texttt{aniruddha.deb.2002@gmail.com}, \texttt{neevahoza@gmail.com}} \quad 
    Sarthak Singla$^{1\dagger}$\quad
    Dinesh Khandelwal$^{2}$ \\
    {\bf Dinesh Garg}$^{2}$ \quad
    {\bf Parag Singla}$^{1}$\\
    $^{1}$Indian Institute of Technology Delhi \quad $^{2}$IBM Research AI\\
}

\newcommand{\B}[0]{$_\text{B}$}

\begin{document}
\maketitle

\begin{abstract}
While {\em forward reasoning} (i.e., find the answer given the question) has been explored extensively in recent literature, backward reasoning is relatively unexplored. 
We examine the {\em backward reasoning} capabilities of LLMs on Math Word Problems (MWPs): given a mathematical question and its answer, with some details omitted from the question, can LLMs effectively retrieve the missing information? 
On modifying three benchmark datasets for this task, to evaluate this task: GSM8k, SVAMP, and MultiArith, we find a significant drop in the accuracy of models on this task compared to forward reasoning across SOTA LLMs (GPT4, GPT3.5, PaLM-2, and LLaMa). Motivated by the fact backward reasoning can be seen as the ``inverse'' of forward reasoning, we propose variations of three different forward reasoning strategies to improve performance. {\it Rephrase} reformulates the given problem into a forward reasoning problem, {\it PAL-Tools} combines the idea of Program-Aided LLMs to produce a set of equations that can be solved by an external solver, and {\em Check your Work} exploits the availability of natural verifier of high accuracy in the forward direction, interleaving solving and verification steps. Finally, realizing that each of our base methods correctly solves a different set of problems, we propose
a novel Bayesian formulation for creating an ensemble over the base methods to further boost the accuracy.
Extensive experimentation demonstrates successive improvement in the performance of LLMs on the backward reasoning task, using our strategies, with our ensemble-based method resulting in significant performance gains compared to the SOTA forward reasoning strategies we adapt.
\end{abstract}

\section{Introduction}
Large language models (LLMs)~\citep{GPT3, GPT4, Palm2} have shown remarkable versatility, excelling in various tasks like sentence completion, question answering, and summarization. They have been successfully applied to mathematical reasoning, specifically in solving {\em Math Word Problems} (MWPs) ~\citep{kushman2014learning, roy2018mapping}, where the goal is to produce the answer given an elementary school-level mathematics question. We refer to this task as {\em Forward Reasoning}. This problem has received significant attention in the recent literature~\citep{lu2022survey}, and specific datasets~\citep{GSM8k,MultiArith,SVAMP} have been proposed as a benchmark for this task. The performance of powerful LLMs such as GPT-4~\cite{GPT4} with techniques such as Chain-of-Thought~\citep{CoT} and Self-Verification~\citep{CodeSelfVerify} on some of these datasets is more than $90\%$~\citep{lu2022survey}. 

We would like to solve a slightly different problem: given an MWP, with one of the numerical quantities omitted from the question, and the answer to the original question, what is the value of the omitted numerical quantity? Yu et al.~\shortcite{MetaMath} refer to this as the problem of {\em Backward Reasoning Problem}, and we would like to examine how effective are LLMs on this task. While this problem of backward reasoning has been studied in the literature in the context of improving the performance of forward reasoning~\citep{SelfVerify}, to the best of our knowledge, there is no existing work that explicitly aims to solve this problem analyzing its hardness and providing solutions thereof. We believe this is an interesting problem because (1) It is a matter of study that even for humans, whether forward reasoning and backward reasoning have different complexities~\citep{ramful2008reversibility, rivera2008pitfalls}, and we would like to ask the same question in the context of LLMs (2) Assuming we establish that backward reasoning is a harder problem, how can we design techniques to improve performance on this task, that specifically exploit the problem structure of backward reasoning and the availability of the forward direction answer? (3) The backward reasoning problem can be seen as a special case of abduction, with a unique answer, and it is interesting to explore this connection since LLMs have not been explored as much for this important class of abductive reasoning problems~\citep{ACR, DeLorean, COLD}.
(4) Enhancing the backward reasoning capabilities of large language models (LLMs) can be highly beneficial in domains such as automated theorem proving~\cite{bibel2013automated, yang2024leandojo}, where solutions are already known and where backward reasoning is essential for automatically generating mathematical proofs.

As an initial analysis, we modify existing benchmark MWP datasets for backward reasoning, and experiment with existing forward reasoning strategies. Interestingly, we observe a significant drop in the LLM accuracy across multiple datasets when working with backward reasoning (refer Table~\ref{tab:model_perf}). We hypothesize this may be due to the specific nature of the task, which makes it harder to solve, or the lack of sufficient data that LLMs have seen during their training compared to forward reasoning. 

We take three different existing forward reasoning strategies and modify them appropriately to make them work effectively for backward reasoning, resulting in (a) Rephrase, based on \cite{SelfVerify} (b) PAL-Tools, based on \cite{PAL} and \cite{Tools}, and (c) Check your work, based on \cite{SelfRefine}. As our final technique, we propose a novel ensemble-based approach combining these base strategies using a Bayesian framework and making use of a {\em forward verifier} whose accuracy is estimated using a hold-out set. Experiments on several benchmark datasets show that we get successive performance improvement using our strategies, with the Bayesian ensemble based approach performing the best, providing 20-30\% points gains compared to SOTA techniques for forward reasoning. We perform additional analysis on the models to explain our results.


Our contributions can be summarized as: (1) we explicitly handle the problem of backward reasoning and identify the performance gap using existing LLM strategies for this task (2) we propose variations of three different existing forward reasoning strategies and a Bayesian ensemble based approach (3) we perform extensive experimentation to demonstrate the efficacy of our models for backward reasoning via LLMs. (4) We perform additional analysis, giving further insights into the performance of the proposed models. We publicly release our code and data\footnote{\url{https://github.com/dair-iitd/fill-in-the-blank-mwp}}.

\section{Related Work}
\begin{figure*}
  \begin{minipage}{\textwidth} 
    \centering
    \includegraphics[width=\textwidth]{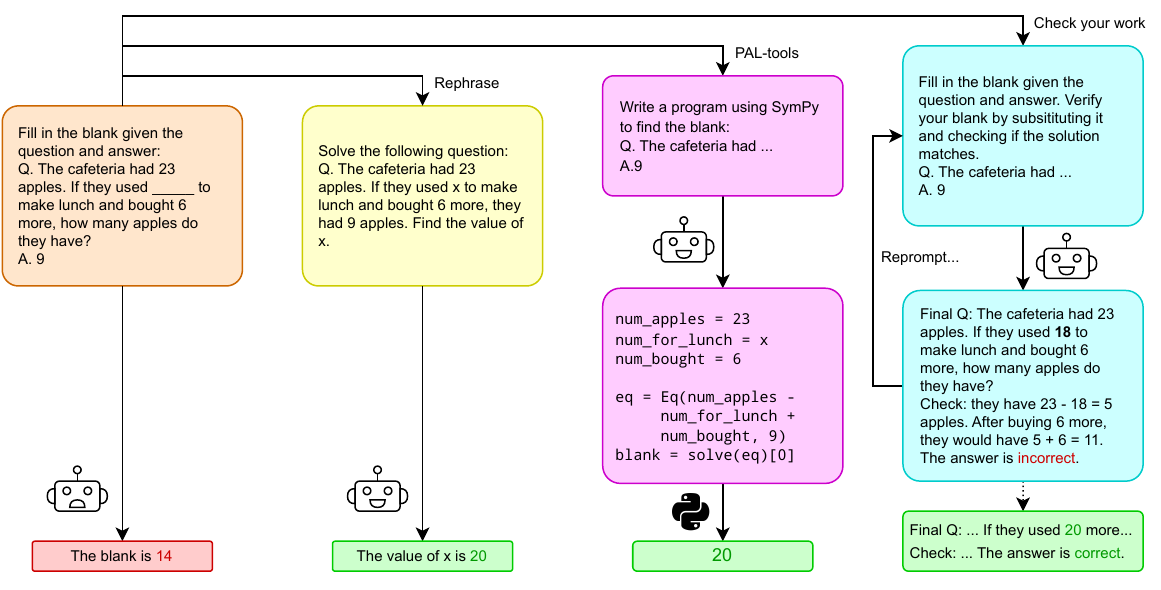}
    \caption{A summary of the prompting techniques we adapt}
    \label{fig:techniques-overview}
  \end{minipage}
\end{figure*}

A mathematical word problem (MWP)~\citep{lu2022survey} consists of a description in natural language that expresses the relation between various entities and quantities, followed by a query for an unknown quantity, as shown in Figure~\ref{fig:techniques-overview}. One can answer the question by representing the relationship between the entities and quantities through a set of equations and then solving these equations. Solving MWPs necessitates a semantic understanding of the natural language description. Initial works~\citep{kushman2014learning, koncel2015parsing, roy2018mapping} to solve MWPs involve parsing the natural language description and utilizing statistical learning techniques to identify suitable templates for generating answers.
Subsequently, following the triumph of sequence-to-sequence (Seq2Seq) neural models~\citep{sutskever2014sequence} in machine translation and other NLP tasks, the encoder-decoder framework~\citep{wang2017deep, ling2017program, li2020graph, shen2021generate, jie2022learning} is employed to directly translate the natural language description in MWPs into equations.

{\bf MWPs:} Recently, the strongest performance on MWPs has been given by large pre-trained language models like GPT-4~\cite{GPT4} and PaLM~\citep{Palm2}. These models leverage the power of few-shot in-context examples and employ prompting methods like CoT~\citep{CoT}, all without requiring any modifications to their parameters. 

\noindent
{\bf Answer Verification:} One class of techniques~\citep{SelfRefine, SelfCorrect} using LLMs involves verifying the answer provided by the Language Model, either using the model itself or external verifiers such as compilers or proof checkers. If the answer is incorrect, the model is re-prompted, optionally with suggestions on improving its output. Other techniques, such as Progressive Hint Prompting \citep{PHP} iteratively pass the model's previous answers to itself as hints. Iterative prompting techniques like \citep{SelfConsistency} do not use a verifier; instead, they sample multiple hypotheses from the model and select the answer using majority voting.

\noindent 
{\bf Backward Reasoning:} Our work can be seen as a special case of abductive reasoning with a unique answer. Abductive reasoning~\citep{ACR, DeLorean, COLD} involves inferring the most plausible out of the several explanations. Prior work on abductive reasoning has focused mostly on text-based reasoning under constraints. In the context of arithmetic reasoning tasks, \citet{SelfVerify} has utilized backward reasoning to enhance forward reasoning accuracy. In contrast, our work addresses backward reasoning as an independent problem. Our primary interest lies in analyzing the inherent complexities of backward reasoning and devising more effective solutions to tackle it.

\section{Task Definition}
\label{sec:task}

\begin{table*}
  \fontsize{9}{11}\selectfont
 \caption{Performance of various models on the MWP backward reasoning task, compared to their accuracies on the forward reasoning task of solving the original problem. Numbers marked $^\dag$ are taken from \cite{PHP}}
\centering
  \begin{tabular}{lrrrrrr}
    \toprule
     & \multicolumn{2}{c}{GSM8k} & \multicolumn{2}{c}{SVAMP} & \multicolumn{2}{c}{MultiArith}  \\
     \cmidrule(lr){2-3}\cmidrule(lr){4-5}\cmidrule(lr){6-7}
    Model & forward & backward & forward & backward & forward & backward \\
    \midrule
    GPT-4          & 92.8 & 38.6 & 90.5$^\dag$  & 43.9  & 97.8$^\dag$  & 54.8 \\
    GPT-3.5-turbo  & 58.4 & 10.8 & 79.1  & 20.4  & 97.0 & 14.5 \\
    PaLM-2         & 60.5 & 15.2 & 73.7  & 11.2  & 95.7 & 6.3  \\
    LLaMa-2-70B    & 37.0 &  6.8 & 70.3  & 20.3  & 89.2 & 11.0    \\
    \bottomrule
  \end{tabular}
  \label{tab:model_perf}
\end{table*}

A forward or the typical Mathematical Word Problem (MWP) consists of a question text $Q_f$, which we call a forward question, and its corresponding answer $A_f$. The forward question is a textual representation of the MWP. It is typically composed of one or more sentences and encompasses various elements, including numbers, operations, and textual information, all represented by tokens within the question. A backward MWP is defined as a tuple $(Q, A_f)$, where $Q$ is obtained from the forward question $Q_f$ by replacing one numerical quantity such as $5$, $3.7$, or ‘half’ with a blank. The goal of solving the backward MWP is to find out the unique value of the numerical quantity that was blanked out using backward reasoning. By backward reasoning, we mean the process of using the provided answer $A_f$ and the context provided by the question $Q$ to deduce the missing numerical quantity to arrive at the given answer. Since there is a unique answer for every question, we measure accuracy on this task by the number of questions on which the model is able to provide the correct numeric value of the blank.

\section{Base Strategies}
\label{sec:approaches}
Table \ref{tab:model_perf} compares the performance of four state-of-the-art (SOTA) language models on forward and backward reasoning tasks by using the 8-shot chain of thought \cite{CoT} prompts. The experiments were conducted using the chain of thought prompts defined in \citet{CoT}. The few-shot examples used in the chain of thought prompts were modified for the backward reasoning task following the procedure described above. A significant drop in backward reasoning accuracy compared to forward reasoning accuracy across all models proves the difficulty of this task for LLMs. Next, we adopt three base approaches for the backward reasoning task, as described below:
    
\paragraph{Rephrase:}
Our first modified SOTA method to tackle the challenging backward reasoning problem involves a problem transformation through rephrasing. This transformation effectively converts the complex backward reasoning task into a more manageable forward reasoning problem. Consequently, we employ the LLM to solve this transformed forward reasoning problem instead of the original and inherently more difficult backward reasoning challenge.
Given a backward MWP $(Q, A_f)$, we ask the language model to produce a rephrased question $R$, which incorporates the forward answer $A_f$ into the question $Q$ and changes the objective of the question from finding the answer $A_f$ to finding the value of the blank. We then ask the language model to solve the rephrased problem $R$ instead of the original backward problem. The verification method used by \cite{SelfVerify} works similarly to this, and they use this task only to improve the accuracy of the forward reasoner. In this prompting strategy, to rephrase the question, the LLM is given in-context examples where the blank is replaced by 'x', $A_f$ is used to change the interrogative part of the question to assertive and the value of 'x' is asked to be found. We define it as {\em algebraic prompt} and have used this for all experiments that include rephrasing the question before solving it. (see the first part of Analysis \ref{sec:other-analysis} for more details).

\paragraph{PAL-Tools:} We modify the Program-aided language model (PAL) \citep{PAL} which writes a Python program to solve the MWP and integrate it with Tools~\cite{Tools} which uses the techniques of framing equations in natural language and calls SymPy to solve them. Neither of the two techniques, i.e., PAL or Tools, does well independently, as shown by our experiments.

\paragraph{Check your Work (CYW):} Inspired from the iterative prompting technique 
\textsc{Self-Refine}~\cite{SelfRefine}, that cycles between refinement and feedback until a stopping criteria is met, 
our approach has the following steps: (1) Generate the answer of $Q$ say $a$. (2) Form a forward problem obtained by substituting the blank with the obtained answer $a$. (3) Check the correctness of $a$ by checking whether the answer to the forward problem matches with the gold answer $A_f$. We repeat this if $a$ is found to be incorrect. Comparing Self-Refine (that we modified to solve the backward task) with Check your Work, Self-Refine uses the LLM's assessment on the backward task as a stopping criteria, whereas in Check your Work, we use the easier problem of forward verification, for deciding when to stop (ref. Appendix \ref{sec:prompts} for prompts). 

\begin{table}[htbp]
 \fontsize{9}{11}\selectfont
\captionsetup{justification=raggedright,singlelinecheck=false}
\caption{Performance of base strategies. LLM is GPT-3.5-Turbo. R: Rephrase. CYW: Check Your Work}
\setlength{\tabcolsep}{3pt}
	\begin{tabular}{p{2.2cm}cccc} 
		\toprule
		Strategy           & Shots & GSM8k\B & SVAMP\B & MultiArith\B \\
		\midrule
		CoT                & 8     & 10.77 & 20.40 & 14.50      \\
		PAL                & 4     & 9.27  & 20.90 & 18.17      \\
		Tools              & 3     & 31.45 & 43.50 & 71.83      \\
            PAL-Tools          & 4     & 37.11 & 42.70 & 80.50      \\
		\midrule
		CoT (R)             & 8     & 36.12 & 37.80 & 71.67      \\
            PAL (R)      & 4  & 21.38 & 37.0 & 55.50  \\
            Tools (R)     & 3  & 41.43  & 48.5  & 73.00  \\
            \midrule
		Self-Refine (R)    & 2     & 40.17 & 49.70 & 77.50      \\
		CYW (R)& 8     & 41.82 & 47.40 & \textbf{84.83} \\
		PAL-Tools (R)      & 4     & \textbf{48.74} & \textbf{51.10} & 84.50 \\	
          \bottomrule
	\end{tabular}
 \label{tab:base-methods}
\end{table}
\newcommand\eq{\mkern1.5mu{=}\mkern1.5mu}
\newcommand\noteq{\mkern1.5mu{\ne}\mkern1.5mu}
\section{A Novel Approach of Ensembling}

\begin{figure*}
    \centering
    \includegraphics[width=\textwidth]{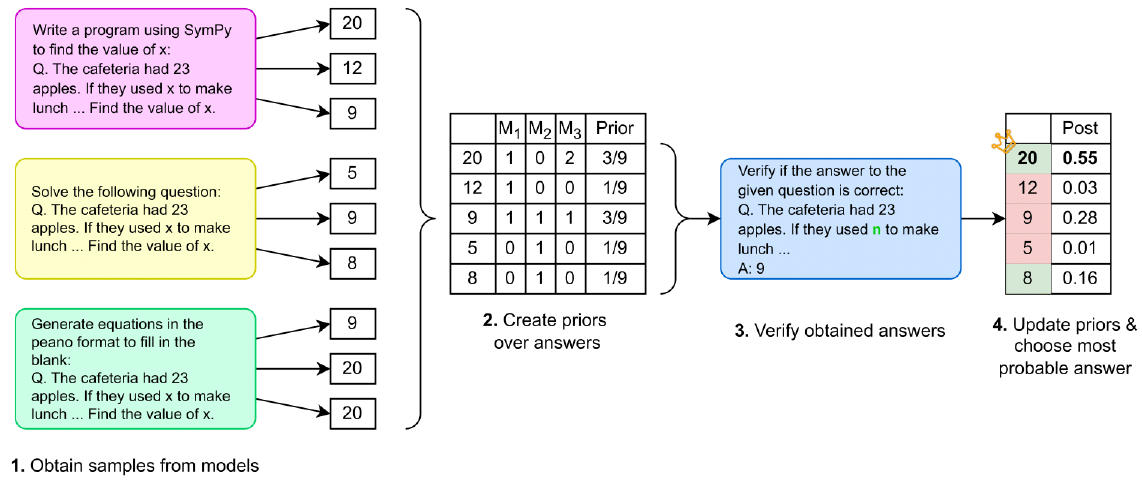}
    \caption{An illustrative example of how the ensembling of base models works together with a verifier.}
    \label{fig:ensemble-overview}
\end{figure*}

We propose a way of ensembling these methods as illustrated by an example in Figure \ref{fig:ensemble-overview}. Assume that we are given a set of models $\{M_1,M_2,\cdots,M_k\}$. Given a model $M_i$, we run the model $M_i$ on the question $Q$ $r$ times, to get a multi-set of answers $\{A_{ij}\}_{j=1}^{r}$. For each unique answer $A$ in this multi-set, we want to estimate the probability of it being correct. We do so by using an LLM as a verifier $V$ in the forward direction, i.e., by substituting the answer in the original question in place of the missing numerical quantity; and solving the forward problem. Given a question $Q$ and an answer $A$, $V$ gives a Boolean output $Z$, which is equal to $1$ if $A$ is the correct answer to the question according to the verifier, and $0$ otherwise. If $A^{*}$ is the gold answer to $Q$, we want the probability of $A$ being correct conditioned on the output of the verifier. From Bayes' rule,

\begin{equation}
    P(A = A^*\ |\ Z, Q) = \frac{X}{X + Y} \label{eq:final}
\end{equation}
where \(X\) and \(Y\) are defined as follows:
\begin{align*}
	X &= {P(Z\ |\ A = A^*, Q)\ P(A = A^*\ |\ Q)}\\
	Y &=  P(Z\ |\ A \neq A^*, Q)\ P(A \neq A^*\ |\ Q)
\end{align*}

We compute the prior $P(A = A^*\ |\ Q)$ as the fraction of times $A$ appears as the answers in the union of the multiset of answers produced by each model for the question $Q:P(A = A^*\ |\ Q)=\sum_{i,j} \mathbbm{1}[A_{ij} \eq a]/kr$. 
\newline

Now, let $P(Z\ |\ A \eq A^*, Q)$ denote the distribution over $V$'s outputs when $A \eq A^*$. Similarly, let $P(Z\ |\ A \noteq A^*, Q)$ denote the distribution over $V$'s outputs when $A \noteq A^*$. We estimate these distributions by computing the accuracy of the verifier on the holdout set $S'$, and supplying a set of answers produced by the $k$ models, each run $r$ times on each $Q \in S'$, along with the gold answer. Thus, we obtain values for Equation (\ref{eq:final}) and select the answer having the highest probability.

\begin{figure}
    \centering
    \includegraphics[width=\columnwidth]{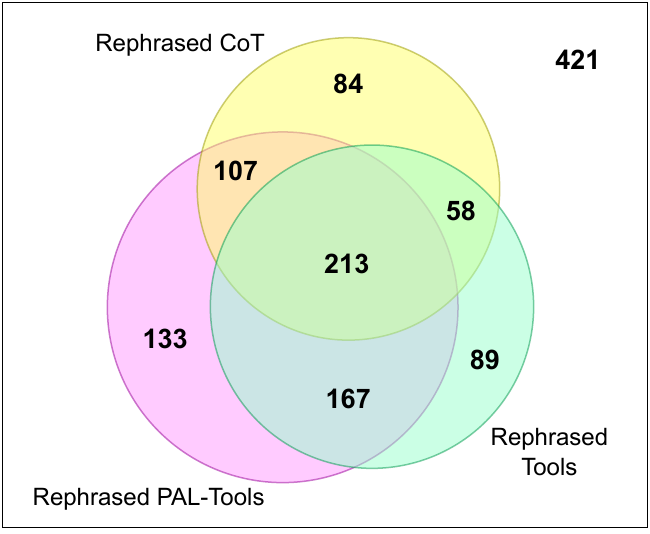}
    \caption{Overlap between problems in GSM8k\B\ that different base techniques can solve}
    \label{fig:ana-venn}
\end{figure}

Note that we could also use the verifier's internal model to estimate $P(Z\ |\ A \eq A^*, Q)$, 
but this may not be well calibrated~\citep{jiang2021can, zhao2021calibrate}, i.e., the model's probability estimates may not accurately reflect the true likelihood of the answer being correct. Therefore, we instead use a holdout set to estimate these probabilities. 
\section{Experiments}
\subsection{Setup}
We start with three forward reasoning datasets: GSM8k~\citep{GSM8k}, MultiArith~\citep{MultiArith}, and SVAMP~\citep{SVAMP}, and transform the examples in these datasets into backward tasks, resulting in the creation of three modified datasets: GSM8k$_\text{B}$, SVAMP$_\text{B}$, and MultiArith$_\text{B}$ (ref. Appendix \ref{sec:dataset} for modification details).
We have experimented with four SOTA LLMs: GPT-4, GPT-3.5-Turbo \citep{GPT4}, PaLM-2 \citep{Palm2} and LLaMa-2 \citep{LLaMa2}.
Further details are in Appendix~\ref{sec:exp-details}.
Prompts and in-context examples for the prompting techniques are taken from their original works. The in-context examples are modified for the backward setting as discussed in Section~\ref{sec:approaches}. Examples of the prompts used are given in Appendix \ref{sec:prompts}. 

\subsection{Results of Base Strategies}

Table~\ref{tab:base-methods} presents the results comparing standard forward reasoning strategies, with their variants that we introduced for backward reasoning~\footnote{we use different number of shots for different models to have roughly the same size of prompts in each case}. Surprisingly, PAL does quite badly on this task, likely because the LLM is not able to construct good programs for the backward reasoning task. 
For rephrasing, we find that models perform better when the rephrased problem has the blank replaced with $x$ compared to the baseline prompt. 
This is because the relationship between the missing value and the equations that models need to frame in order to solve the forward problem is explicit. Also, for the baseline prompt, the model requires inferring the relationship between the forward answer and the equations they need to frame to obtain it, which may introduce ambiguity and reduce accuracy.
We see that Rephrase also helps the other standard techniques in all cases, and we use that along with our remaining models. 

CYW does better than Self-Refine on two of the datasets, gaining significantly on MultiArith$_\text{B}$. This points to the efficacy of using LLM as a forward verifier.
The best-performing model is PAL+Tools, significantly improving accuracy over both PAL and Tools when run independently. We hypothesize the combination allows the model to retain the advantages of the programmatic way of formulation, while the backward reasoning is handled effectively by calling the external solver as in Tools.

\begin{table}
  \centering
    \fontsize{9}{11}\selectfont
    \caption{Ensembling results. LLM: GPT-3.5-Turbo}
    \begin{tabularx}{\columnwidth}{p{2.2cm}ccc}

      \toprule
           & GSM8k$_\text{B}^\dag$ & SVAMP$_\text{B}^\dag$ & MultiArith$_\text{B}^\dag$ \\
      \midrule
      CoT (R)            &  35.67 & 37.78 & 69.60 \\
      Tools (R)          &  41.81 & 48.11 & 72.00 \\
      PAL-Tools (R)      &  48.55 & 45.00 & 81.50 \\
      Majority Voting    &  58.28 & 59.07 & 92.00 \\ 
      Ensemble           &  \textbf{65.33} & \textbf{66.67} & \textbf{92.60}\\
      \bottomrule
    \end{tabularx}
    
    \label{tab:ensembling}
    
\end{table}

\subsection{Ensembling}
The models we included in the ensemble are rephrased versions of three of our strongest single-prompt models: CoT, Tools, 
and PAL-Tools. We use a temperature of $0.5$ for sampling to generate $3$ answers per question with each model. With $3$ different models, we generate a total of $9$ answers per question. The verifier is the LLM used in each of these base strategies. We select $100$ examples from the 
datasets as holdout sets to compute an estimate of the verifier accuracy. We evaluate all the models on the non-holdout set, 
which is denoted with a $^\dag$ symbol in Table \ref{tab:ensembling} to show the results. Clearly, we see that our 
ensembling technique results in significant gains ($10$-$20\%$) on all the datasets, compared to the best-performing PAL-Tools which is the best-performing base model (ref. Table~\ref{tab:base-methods}). We also compare our ensemble-based method with plain majority voting. Our approach results in close to $7\%$ gain compared to vanilla majority voting pointing on two of the datasets to the efficacy of our Bayesian ensembling via a forward verifier.
We observe that the accuracy on backward MWP via ensembling surpasses the forward accuracy of CoT by up to $6\%$. We also show that ensembling improves performance compared to majority voting, it shows the significance of using a verifier for selecting the correct answer compared to generating multiple answers via LLM. 

On analyzing the base methods, as shown in figure \ref{fig:ana-venn}, we find that there are a significant number of problems that are solved correctly by one of the models but not by others. This shows how ensembling exploits the strenghts of individual models to provide significant boost in overall performance.
Ensembling exploits the fact that the task of verifying is easier for LLMs than solving the backward MWP. 
\section{Analysis}
\label{sec:other-analysis}
\begin{description}[leftmargin=0pt]
\item[How much does rephrasing help?] Since rephrasing is a strategy that can be applied across multiple techniques, we analyze the extent of accuracy gains obtained via rephrasing by applying it independently to the base techniques: CoT, PAL, and Tools. The results are shown in Table \ref{tab:base-methods}. Rephrasing improves the accuracy of every technique that it is applied to. We see larger gains with rephrasing in weaker methods, such as CoT. We also see that rephrasing has higher gains in datasets where the problems are harder, such as in GSM8k compared to SVAMP. We also tried to check whether the nature of the prompt affected the performance of rephrasing. We tried using a prompt where unlike the {\em algebraic prompt}, instead of introducing an 'x' and asking to find its value, we gave in-context examples that convert the problem to be worded similar to a forward reasoning task MWP. We call this a {\em linguistic prompt} In the experiment of table \ref{tab:rephrasing}, we show how our specific prompts improve the performance of CoT. And we demonstrate the advantage of explicitly naming the value to be found as 'x'.

\begin{table}
\centering
  \fontsize{9}{11}\selectfont
 \caption{Improvements in accuracy with rephrasing strategies. LP: Linguistic prompt rephrase. AP: Algebraic prompt rephrase}
 \label{tab:rephrasing}
  \begin{tabularx}{\columnwidth}{p{1.5cm}lcccc}
    \toprule
    Strategy                      & Shots & GSM8k\B & SVAMP\B & MultiArith\B \\
    \midrule
    CoT                           & 8 & 10.77 & 20.40  & 14.50 \\

    CoT  (LR)         & 8 & 19.65 & 32.60  & 40.50 \\

    CoT  (AR)          & 8 & 36.12 & 37.80  & 71.67 \\
    \bottomrule
  \end{tabularx}
  \label{tab:rephrase-versions}
\end{table}

\item[Is verifying easier than solving?] In the third step of the ensembling method, we try to verify whether the blank provided is correct by solving the resulting forward problem after substituting the blank. There are two settings in which we can verify this: 1) We ask the model to solve this new question and compare whether the answer obtained is the same as the original answer $a$.
2) We give the original answer $a$ to the model and ask it to check whether it is the answer obtained for the new question.

To find which method is better at correctly verifying the blank, we check the accuracy of GPT-3.5-turbo on GSM8k in setting 2. In the first pass, we provide the correct blank and in the second pass, we provide an incorrect blank formed by multiplying $z \in \{2 \ldots 10\}$ with the correct blank. The confusion matrix obtained is shown in Table \ref{tab:verif_conf_mat}. It is observed that the accuracy of setting 2 is higher than the forward reasoning accuracy of GPT-3.5-turbo. Hence, we use that as the verification method for ensembling.

\begin{table}
    \centering
    \begin{minipage}{0.4\textwidth}
        \centering
      \fontsize{9}{11}\selectfont
      \begin{tabular}{lrr}
        \toprule
         & \multicolumn{2}{c}{model} \\ \cmidrule(ll){2-3} 
         actual & positive & negative \\
        \midrule
        positive  & 75.94 & 24.05 \\
        negative  & 7.39 & 92.61  \\
        \bottomrule
      \end{tabular}
      \caption{Confusion matrix for verifying problems and their solutions on GSM8k\B, normalized across rows}
      \label{tab:verif_conf_mat}
    
    \end{minipage}
    \hspace{1cm}
    \begin{minipage}{0.45\textwidth}
        \centering
        \includegraphics[width=\columnwidth]{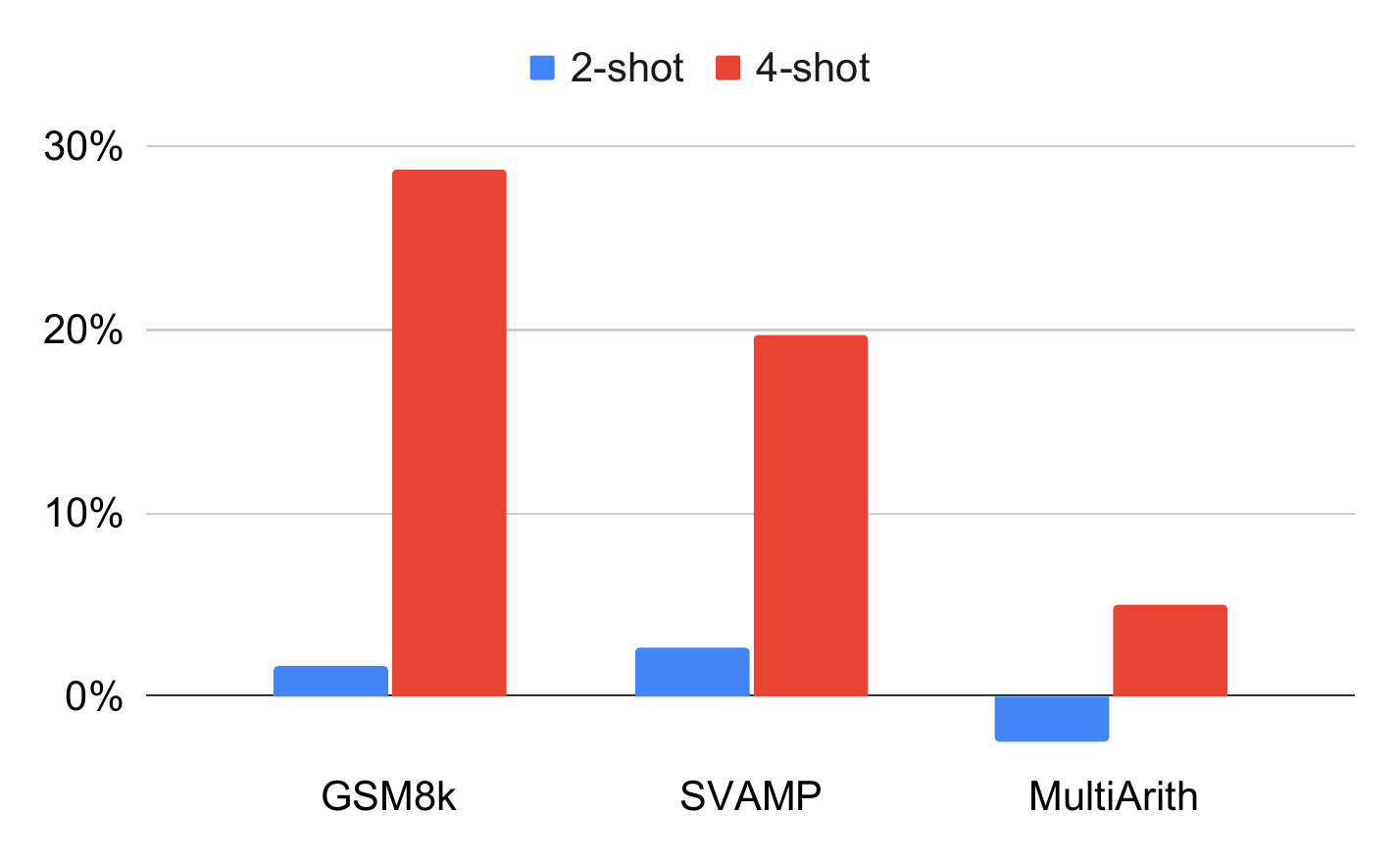}
        \caption{Relative performance increase rephrasing brings to PAL-tools when 4 shots are used compared to 2.}
        \label{fig:pal_tools_2_vs_4}
    \end{minipage}
\end{table}

\item[Does the verifier assist in ensembling?] We compare the accuracies obtained by using majority voting with and without the verifier. We find that using a verifier improves the accuracy on GSM8k$_\text{B}^\dag$ and SVAMP$_\text{B}^\dag$ by 7\% and on MultiArith$_\text{B}^\dag$ by 0.6\%. Since the verifier has a higher accuracy than any of the models we consider, its inclusion inevitably increases the accuracy of any set of methods we choose. Even if we use a noisy verifier, updating our priors based on its results using Bayes' rule ensures that the priors are not changed significantly.

\item[Do some prompts subsume others?] Let prompts $M_1, M_2, M_3$ be able to solve problems $\mathcal{D}_1, \mathcal{D}_2, \mathcal{D}_3$ respectively, where $\mathcal{D}_i \subseteq D$. If we choose to ensemble these prompts together, then $|\bigcup_j \mathcal{D}_j| > |\mathcal{D}_i|$ for the ensemble to do better. Figure \ref{fig:ana-venn} gives an overview of the subsets of GSM8k that the three prompts, namely Rephrased Chain of Thought, Rephrased PAL-Tools, and Rephrased Tools can solve in a single try. We see that even though there is significant overlap between the prompts, the probability of any one of them giving the right answer is $66.9\%$, provided we sample from each prompt once. The Venn diagram also shows the subsets of problems different prompts cover are quite disjoint in nature. No prompt can solve all the problems that another prompt can solve.




\item[Can every blank's answer be determined?] There may be cases where the blank does not directly contribute to the answer or is irrelevant. In such a case, inferring the value of the blank is not possible given the answer. Even though \cite{GSM8k} claim that less than two percent of problems have breaking errors, We sample 50 random examples from GSM8k\B that our strongest model solves incorrectly and find that no such problems in the sample we analyse, leading us to believe that the probability of such problems existing in our dataset is little to none.

\item[Do all blanks need answers to be solved?] In the 50 examples we analyse above, there are 10 examples where the value of the blank can be obtained simply from reading the question, as the question makes implicit assumptions or provides further information that can be used to fill in the blank. Two examples are presented in table \ref{fig:bad_examples}. It is surprising that even our strongest model is unable to find the answer to such questions, either as a consequence of its poor reasoning abilities or because we make the dependency between requiring the answer to fill in the blank explicit.

\begin{table}
  \fontsize{9}{11}\selectfont
  \centering
        \caption{ Examples of questions in GSM8k\B\ that don't require the answer to find the value of the blank}

        \begin{tabular}{p{0.9\columnwidth}}
            \toprule
             Carla just gave birth to identical octuplets. She dresses 3/4 of them in purple and \_\_\_\_\_ in blue. If all the blue wearers and 1/3 of the purple wearers also wear bows, what is the percentage chance a baby wearing a bow is wearing purple? \\
             \midrule
             Ian has a board that is 40 feet long. He decides to make a cut so he can have \_\_\_\_\_ pieces. The longer piece is 4 times longer than the shorter piece. How long is the longer piece? \\
             \bottomrule
        \end{tabular}
        \label{fig:bad_examples}
\end{table}

\end{description}
\section{Going beyond masking of a single numeric quantity}
\label{sec:phrase-mask}

As a preliminary study, we aim to extend the task by forming tuples $(Q, A_f)$ by masking a phrase instead of a numeric quantity. A phrase is defined as the contiguous set of words between two connectives such as [‘and’, ‘,’ , ‘.’], and from the forward question $Q_f$, we replace the phrase containing the second occurrence of any numeric quantity with a blank. This choice is made to make the task similar to abductive reasoning~\citep{ACR, DeLorean, COLD} for story completion
. We define this as a phrase-masked backward reasoning task. Note that in this setting, there can be multiple correct answers, which may differ from the originally masked phrase. To verify the correctness of the generated phrases, the task requires hand-evaluation of the results produced by various techniques. 

As a baseline, we prompt the LLM to fill in the masked phrase given the rest of the question and the final answer to the original question using CoT. We observed experimentally that asking LLM to divide the task to first guessing the missing phrase with an 'x' in it and then solving it similar to a single numeric quantity masked backward reasoning task, performed better than finding the entire phrase altogether. We modified our methods to include a prior step to guess in this manner. For the hand-evaluation of the method, we replace the blank phrase in $Q$ with the predicted phrase and replace the value of 'x' in it with the value generated by the respective method. We then manually solve the resulting question and if its answer is the same as $A_f$, we mark that generation as correct.
We performed the experiments on a $100$-sized subset of the GSM8k dataset.  We used GPT-3.5-Turbo with the same experimental setup as before. As it can be seen in table \ref{tab:phrase-masked}, our methods improve performance for this extension of the task.

\begin{table}
\centering
  \fontsize{9}{11}\selectfont
 \caption{Performance on 100 examples for phrase-masked backward reasoning task. Note: The accuracy of CoT on the forward task of these 100 examples is 80\%. Ensembling uses 3 rephrased methods: CoT, Tools, and PAL-Tools 
 *The number of shots for each strategy used in ensembling matches the number used when the strategy is applied individually.}
  \begin{tabularx}{\columnwidth}{lcl}
    \toprule
    Strategy                & Shots  & Accuracy \\
    \midrule
     CoT                  & 8 & 22  \\
     Check Your Work    & 8 & 27 \\
     CoT Rephrase        & 8   & 37 \\
     Rephrased Tools       & 3  & 34 \\
     Rephrased PAL-Tools    & 4   & 38 \\
     Ensembling            & * & \textbf{61.16} \\
    
    \bottomrule
  \end{tabularx}
  \label{tab:phrase-masked}
\end{table}


\begin{description}[leftmargin=0pt]

\item[Esembling:] As the multiset of answers for ensembling consists of phrases instead of numbers, almost all answers have minor differences in wording, capitalization, etc. 
Thus, as we use exact phrase matching to count the frequency of an answer, even semantically identical answers get counted as different. Additionally, the probability of the verifier giving output $1$ for a correct answer is independent of the question and answer. When these factors are applied in Equation~\ref{eq:final}, they result in a set $S_m$ where each answer marked as correct by the verifier, has an equal probability of being correct. This differs from number masking, where typically only a single answer has the highest probability of being correct. Therefore, we hand-evaluate all the answers in $S_m$, and use the fraction of correct answers in $S_m$ as the probability that an answer sampled from $S_m$ is correct.

\item[Caveat:]
On analyzing, we found that for a few generations, the predicted phrase trivially requires the LLM to directly guess the quantity $A_f$ as 'x' and then it solves for 'x' (for now, we have marked it correct in our evaluation); 
thus making it technically correct without actually solving the backward reasoning task. An example of such generation is depicted in table \ref{tab:trivial_phrase}. 
Tricky MWPs include questions with extra information, which creates a spectrum depending on how many lines of questions are needed to find the answer. The phrases generated by the LLM achieve their goal but may simplify MWPs to be solved with just one line of information. Table~\ref{tab:number_trivial_phrase} shows the counts of such correctly generated phrases. For Ensembling, in case of multiple correct answers, we report the fraction of answers that are trivially correct as described above. We will try other methods in the future to improve on this aspect.


\begin{table}
  \fontsize{9}{11}\selectfont
  \centering
        \caption{Example of an LLM generation that frames the phrase as a guess of $A_f$}

        \begin{tabular}{p{0.9\columnwidth}}
            \toprule
             \textbf{Phrase-masked question:} \\  
             $Q$: A football team played 22 games. \_\_\_\_\_. How many did they win? \\
             $A_f$: 15\\
             \midrule
             \textbf{The phrase that was masked:} \\ 
             They won 8 more than they lost \\
             \midrule
             \textbf{LLM generation (by CoT rephrase method):} \\
             Guess: They won x games \\
Rephrased: A football team played 22 games. They won x games. They won 15 games. Find the value of x. \\
Answer: The football team played 22 games. If they won x games, then they won x out of 22 games. So the answer is x = 15. \\
            \midrule
            \textbf{Phrase that was finally extracted from LLM:} \\
            They won 15 games
            \\
             \bottomrule
        \end{tabular}
        \label{tab:trivial_phrase}
\end{table}

\begin{table}
\centering
  \fontsize{9}{11}\selectfont
 
 \caption{Number of questions in the 100-sized subset that are like the example in table \ref{tab:trivial_phrase}. }
  \begin{tabularx}{\columnwidth}{lr}
    \toprule
    Strategy               & Number of questions \\
    \midrule
     CoT                  & 14.00   \\
     Check Your Work    & 9.00 \\
     CoT Rephrase        & 6.00   \\
     Rephrased Tools       & 2.00  \\
     Rephrased PAL-Tools    & 1.00   \\
     Ensembling            & 5.28  \\
    
    \bottomrule
  \end{tabularx}
  \label{tab:number_trivial_phrase}
\end{table}

\end{description}
\section{Conclusion and Future Work}
We consider the problem of backward reasoning in MWPs. We show that existing forward reasoning strategies do not work well off-the-shelf for the forward problem, and propose 3 different variations of existing techniques for the backward task. We also propose a novel Bayesian ensemble based approach to further improve the accuracy. Experiments demonstrate the efficacy of our approach compared to forward reasoning strategies. 
Finally, we analyze the fallacies and pitfalls of each of these techniques and show areas for future improvements. Our method of Bayesian ensembling can also be extended to other tasks of backward reasoning. If there is a setting where the verification or the forward reasoning task is easier compared to the backward, and there is an existing set of methods to solve the backward task in different ways; our method can potentially be used in such settings. 
Our future work includes extending our technique to other backward reasoning datasets, including those derived from explicit abductive reasoning tasks.

\section{Limitations}
Limitations of our approach include (a) It has only been tested on the MWPs. (b) It requires a (small) hold-out set to estimate the accuracy of the verifier in the ensemble.

\section*{Acknowledgements}
This work was supported by an IBM AI Horizons Network (AIHN) grant and IBM SUR Awards. We thank IIT Delhi
HPC facility\footnote{https://supercomputing.iitd.ac.in/}, IBM cloud facility, and IBM Cognitive Computing Cluster (CCC) for computational
resources. We thank anonymous reviewers for their insightful comments that helped in further
improving our paper. 
Any opinions, findings, conclusions or recommendations expressed in this paper are
those of the authors and do not necessarily reflect the views or official policies, either expressed or
implied, of the funding agencies.


\bibliography{custom}
\bibliographystyle{acl_natbib}

\appendix
\titleformat{\section}[block]{\large\bfseries}{\appendixname~\thesection: }{0em}{}
\onecolumn

\section{Dataset}
\label{sec:dataset}

We consider three datasets of interest: GSM8K \citep{GSM8k}, MultiArith \citep{MultiArith}, and SVAMP \citep{SVAMP}. All these datasets consist of grade-school arithmetic word problems along with their answers. 

\subsection{Generation Methodology}

Given a source forward dataset 
$$D = \{ (Q_i, A_i)_{i=1}^n\ |\ Q_i \in \Sigma^*, A_i \in \mathbb{R} \}$$
we present a method to create a backward dataset 
\begin{align*}
D'_k &= \{ (Q_i', A_i, (B_i^0, \ldots, B_i^k))_{i=1}^{n} \ | \ Q'_i \in \Sigma^*, A_i, B_i^j \in \mathbb{R} \}
\end{align*}
To convert $Q_i$ (Source question) to $Q_i'$ (blanked out question) and extract blanks $B_i^0 \ldots B_i^k$, we split $Q_i$ into its constituent tokens based on a delimiter, usually space. We then consider all numeric tokens, which are defined as tokens that encode a number. Numeric tokens may be alphanumeric, such as \$42, 80\% or 3.14, or they may be alphabetic, such as three, twice, or half. Using this heuristic for numeric tokens, we ignore the first numeric token and extract the next $k$ tokens sequentially. If we are unable to extract $k$ tokens, then we skip that question and answer pair. It is worth noting that for the datasets we use, $k = 1$, that is we only consider the problem of backwardly inferring one missing number in the question, given the answer. Solving the $n > 1$ case would require first checking if a unique solution exists and is a topic for future work.

The reason we choose to blank out only numeric tokens rather than an entire phrase or sentence is to make the task of validation easier. An alternative that was explored was phrase masking. However, phrase masking would lead to generations that would not be verifiable with perfect accuracy, and multiple possible generations for each question. The benefit of number masking is that quantities can be compared to each other without loss of accuracy, and every question-answer pair has a unique blank.

\subsection{Generation Results}

Using the above method, we were able to convert 1272 of the 1319 question and answer pairs in GSM8k to backward reasoning problems, and all 1000 and 600 pairs in SVAMP and MultiArith respectively. All of these are obtained from the respective test splits of the original datasets.

\subsection{Dataset Examples}
Some examples of the datasets under consideration are shown in Table \ref{tab:dataset_examples}. Note that for GSM8k, the original dataset contains 1319 sample problems but our dataset generation method for the backward task filters out 47 of them. For comparability with the backward task, we have used the 1272 common examples of this dataset. 

\begin{table}[H]
    \centering
    \fontsize{9}{11}\selectfont
    \begin{tabular}{p{0.1\linewidth} p{0.05\linewidth} p{0.7\linewidth}}
        \toprule
         Dataset & \textit{N} & Example \\
         \midrule
         GSM8k &1272\textsuperscript{\rm *} & Kylar went to the store to buy glasses for his new apartment. One glass costs \$5, but every second glass costs only 60\% of the price. Kylar wants to buy 16 glasses. How much does he need to pay for them? \\
         \midrule
         GSM8k\B &1272 & Q : Kylar went to the store to buy glasses for his new apartment. One glass costs \$5, but every second glass costs only \_\_\_\_\_\% of the price. Kylar wants to buy 16 glasses. How much does he need to pay for them? \\ & & A : 64 \\
         \midrule
         SVAMP &1000 & 28 children were riding on the bus. At the bus stop 82 children got on the bus while some got off the bus. Then there were 30 children altogether on the bus. How many more children got on the bus than those that got off? \\
         \midrule
         SVAMP\B &1000 & Q : 28 children were riding on the bus. At the bus stop, \_\_\_\_\_ children got on the bus while some got off the bus. Then there were 30 children altogether on the bus. How many more children got on the bus than those that got off?" \\ & & A : 2 \\
         \midrule
         MultiArith &600 & Lana picked 36 tulips and 37 roses to make flower bouquets. If she only used 70 of the flowers though, how many extra flowers did Lana pick? \\
         \midrule
         MultiArith\B &600 & Q : Lana picked 36 tulips and \_\_\_\_\_ roses to make flower bouquets. If she only used 70 of the flowers though, how many extra flowers did Lana pick? \\ & & A : 3 \\
         \bottomrule
    \end{tabular}
    \caption{Sample questions from the datasets we consider}
    \label{tab:dataset_examples}
\end{table}

\section{Experiment Reproducibility Details}
\label{sec:exp-details}
For all experiments involving closed-source LLMs, (i.e. GPT-4, GPT-3.5-turbo, and PaLM-2), we utilized the respective model APIs (OpenAI, Google Bard). For experiments using LLaMa-2, we use the 70-billion-parameter model quantized to 4-bit using GPTQ~\citep{GPTQ}. Inference was performed on two 40GB NVIDIA A100 GPUs, on a High Performance Computing cluster node with 8 cores and 16 GB of RAM allocated to the job.
For all models, the temperature was set to 0.5 and the maximum number of tokens to generate was limited to 1024.
In all the tables showing results, the accuracy values are obtained by taking the mean across all examples of the dataset, \emph{in a single run of the mentioned method}.

\section{Prompts}
\label{sec:prompts}
We construct prompts by changing the original examples of the papers we consider to solve the backward task. We show one to two in-context examples of each prompt. The remaining examples may be seen in our code. 

\begin{figure}[H]
\centering
\begin{minipage}{\columnwidth}
\begin{lstlisting}
Rephrase the given blanked question and answer pairs and then find the solution to the rephrased question. Give your answer as either a number or a decimal (no fractions). Follow the format specified in the examples below:

Q: There are 15 trees in the grove. Grove workers will plant _____ trees in the grove today. After they are done, how many trees would be there?
A: 21 
Rephrased: There are 15 trees in the grove. Grove workers will plant some trees in the grove today. After they are done, there would be 21 trees. Find the number of trees planted.
Answer: There are 15 trees originally, Then there were 21 trees after some more were planted. So there must have been 21 - 15 = 6 trees. The answer is 6.

Q: If there are 3 cars in the parking lot and _____ more cars arrive, how many cars are in the parking lot?
A: 5
Rephrased: If there are 3 cars in the parking lot and some more cars arrive, there are 5 cars in the parking lot. Find the number of cars that arrived.
Answer: There are originally 3 cars. There are 5 cars after some more cars arrive. 5 - 3 = 2, so 2 cars arrived. The answer is 2.

...

Q: {{question}}
A: {{answer}}
Rephrased:

\end{lstlisting}
\end{minipage}
\caption{Rephrasing with Linguistic prompt}
\label{prompt:rephrase-linguistic}
\end{figure}

\begin{figure}[H]
\centering
\begin{minipage}{\columnwidth}
\begin{lstlisting}
Rephrase the given blanked question and answer pairs and then find the solution to the rephrased question. Give your answer as either a number or a decimal (no fractions). Follow the format specified in the examples below:

Q: There are 15 trees in the grove. Grove workers will plant _____ trees in the grove today. After they are done, how many trees would be there?
A: 21 
Rephrased: There are 15 trees in the grove. Grove workers will plant x trees in the grove today. After they are done, there would be 21 trees. Find the value of x.
Answer: There are 15 trees originally, Then there were 21 trees after some more were planted. So there must have been x = 21 - 15 = 6 trees. The answer is 6.

Q: If there are 3 cars in the parking lot and _____ more cars arrive, how many cars are in the parking lot?
A: 5
Rephrased: If there are 3 cars in the parking lot and x more cars arrive, there are 5 cars in the parking lot. Find the value of x.
Answer: There are originally 3 cars. x more cars arrive. 3 + x = 5, so x = 5 - 3 = 2. The answer is 2.

...

Q: {{question}}
A: {{answer}}
Rephrased:

\end{lstlisting}
\end{minipage}
\caption{Rephrasing with  Algebraic prompt}
\label{prompt:rephrase}
\end{figure}

\begin{figure}[H]
\centering
\begin{minipage}{\columnwidth}
\begin{lstlisting}
You are given a math question with a blank value and an answer. Solve it step b step to find the value of blank. Strictly follow the format given in the examples below.

Question: Ben has four boxes with ten basketball cards in each box. Ben received ______ cards from his mother. If he gives 58 cards to his classmates, how many cards does he has left?
Answer: 22

Peano solution:


Let a be number of boxes [[var a]]. We have [[eq a = 4]].
Let b be number of cards in each box [[var b]]. We have [[eq b = 10]].
Let c be number of cards Ben initially has [[var c]]. We have [[eq c = a * b]].
Let d be cards received from mother [[var d]].
Let e be cards given to classmates [[var e]]. We have [[eq e = 58]].
Let f be cards left [[var f]]. From given Answer, we have [[eq f = 22]]. 
We have [[eq d = f + e - c]]
The answer is the value of d [[answer d]].




Question: Natalia sold _____ clips to her friends in April, and then she sold half as many clips in May. How many clips did Natalia sell altogether in April and May?
Answer: 72

Peano solution:


Let a be number of clips Natalia sold in April [[var a]].
So number of clips Natalia sold in May are half of a.
Let b be number of clips sold altogether [[var b]]. From given Answer, we have [[eq b = 72]].
We have [[eq a = b / (1 + 1/2)]]
The answer is the value of a [[answer a]].

...

Q: {{question}}
A: {{answer}}

Peano solution:

\end{lstlisting}
\end{minipage}
\caption{Tools}
\label{prompt:Tools}
\end{figure}

\begin{figure}[H]
\centering
\begin{minipage}{\columnwidth}
\begin{lstlisting}
Rephrase the given blanked question and answer pairs and then solve it step b step to find the value of blank. Strictly follow the format given in the examples below.

Question: Ben has four boxes with ten basketball cards in each box. Ben received ______ cards from his mother. If he gives 58 cards to his classmates, how many cards does he has left?
Answer: 22

Rephrased: Ben has four boxes with ten basketball cards in each box. Ben received x cards from his mother. If he gives 58 cards to his classmates, he has 22 cards left. Find the value of x.
Peano solution:


Let a be number of boxes [[var a]]. We have [[eq a = 4]].
Let b be number of cards in each box [[var b]]. We have [[eq b = 10]].
Let c be number of cards Ben initially has [[var c]]. We have [[eq c = a * b]].
Let x be cards received from mother [[var x]].
Let d be total cards with Ben [[var d]]. We have [[eq d = c + x]]
Let e be cards given to classmates [[var e]]. We have [[eq e = 58]].
Let f be cards left [[var f]]. From given Answer, we have [[eq f = 22]]. 
As cards left are also total cards - cards given to classmates, we have [[eq f = d - e]]
The answer will be the value of x [[answer x]].





Question: Natalia sold _____ clips to her friends in April, and then she sold half as many clips in May. How many clips did Natalia sell altogether in April and May?
Answer: 72

Rephrased: Natalia sold clips to x of her friends in April, and then she sold half as many clips in May. Find the value of x such that she sold a total of 72 clips altogether in April and May.
Peano solution:


Let x be number of clips Natalia sold in April [[var x]]
Let a be number of clips Natalia sold in May [[var a]]. We have [[eq a = x / 2]].
Let b be number of clips sold altogether [[var b]]. From given Answer, we have [[eq b = 72]].
As clips sold altogether are also the sum of clips sold in April and May, we have [[eq b = x + a ]]
The answer will be the value of x [[answer x]].

...

Q: {{question}}
A: {{answer}}

Rephrased:
\end{lstlisting}
\end{minipage}
\caption{Tools with Rephrasing}
\label{prompt:rephrase_tools}
\end{figure}

\begin{figure}[H]
\centering
\begin{minipage}{\columnwidth}
\begin{lstlisting}
Rephrase the given blanked question and answer pairs and then find the solution to the rephrased question. Write a python function that finds the value of x by solving step by step. Make sure you name your method finding_x. A few examples are given below.:

Question: Ben has four boxes with ten basketball cards in each box. Ben received ______ cards from his mother. If he gives 58 cards to his classmates, how many cards does he has left?
Answer: 22
Rephrased: Ben has four boxes with ten basketball cards in each box. Ben received x cards from his mother. He gives 58 cards to his classmates. He has 22 cards left.
Program:
```python
def finding_x():
    num_boxes = 4
    cards_per_box = 10
    # cards_received_from_mother = x - This line is commented because x is unknown
    # hence the variable cards_received_from_mother can't be used in R.H.S. of any calculation
    cards_given_to_classmates = 58
    cards_left = 22
    cards_in_boxes = num_boxes * cards_per_box
    total_cards_before_given_to_classmates = cards_given_to_classmates + cards_left
    
    cards_received_from_mother = total_cards_before_given_to_classmates - cards_in_boxes 
    return cards_received_from_mother
```

Question: Olivia has $23. She bought _____ bagels for $3 each. How much money does she have left?
Answer: 8 
Rephrased: Olivia has $23. She bought x bagels for $3 each. She has $8 left. Find the value of x.
Program:
```python
def finding_x():
    money_initial = 23
    # num_of_bagels = x - This line is commented because x is unknown
    # hence the variable num_of_bagels can't be used in R.H.S. of any calculation
    bagel_cost = 3
    money_left = 8
    money_spent = money_initial - money_left
    
    num_of_bagels = money_spent / bagel_cost
    return num_of_bagels
```

Question: {{question}}
Answer: {{answer}}
Rephrased:
\end{lstlisting}
\end{minipage}
\caption{PAL with Rephrasing}
\label{prompt:rephrase_pal}
\end{figure}

\begin{figure}[H]
\centering
\begin{minipage}{\columnwidth}
\begin{lstlisting}
Rephrase the given blanked question and answer pairs and then write a python function called solution() to find the value of x in the rephrased question. Return the value of x. You may assume the neccessary libraries are imported. Strictly follow the format given in the examples below, as the method will be executed with the same name.

Q: Ben has four boxes with ten basketball cards in each box. Ben received _____ cards from his mother. If he gives 58 cards to his classmates, how many cards does he has left?
A: 22
Rephrased: Ben has four boxes with ten basketball cards in each box. Ben received x cards from his mother. He gives 58 cards to his classmates. He has 22 cards left. Find the value of x.
Program:
```python
def solution():
    num_boxes = 4
    cards_per_box = 10
    total_cards_in_boxes = num_boxes * cards_per_box
    cards_from_mother = x
    cards_given_to_classmates = 58
    cards_left = 22
    
    equation = Eq(cards_from_mother + total_cards_in_boxes, cards_given_to_classmates + cards_left)
    blank = solve(equation)[0]

    return blank
```

Q: Natalia sold _____ clips to  her friends in April, and then she sold half as many clips in May. How many clips did Natalia sell altogether in April and May?
A: 72
Rephrased: Natalia sold x clips to her friends in April, and then she sold half as many clips in May. Natalia sells 72 clips altogether in April and May. Find the value of x.
Program:
```python
def solution():
    april_clips = x
    may_clips = april_clips / 2
    total_clips = 72
    
    equation = Eq(april_clips + may_clips, total_clips)
    blank = solve(equation)[0]

    return blank
```

...

Q: {{question}}
A: {{answer}}
Rephrased:
\end{lstlisting}
\end{minipage}
\caption{PAL-Tools with Rephrasing}
\label{prompt:rephrase_pal_tools}
\end{figure}

\begin{figure}[H]
\centering
\begin{minipage}{\columnwidth}
\begin{lstlisting}
Fill in the blank given the question and answer examples below. Give your answer as either a number or a decimal (no fractions). Check your work by substituting your answer in the blank, solving the question and comparing to the original answer. Follow the format specified in the examples below:

Q: There are 15 trees in the grove. Grove workers will plant _____ trees in the grove today. After they are done, how many trees would be there?
A: 21 
Rephrased: There are 15 trees in the grove. Grove workers will plant x trees in the grove today. After they are done, there would be 21 trees. Find the value of x.
Answer: There are 15 trees originally, Then there were 21 trees after some more were planted. So there must have been x = 21 - 15 = 6 trees. The answer is 6.
Final question: There are 15 trees in the grove. Grove workers will plant 6 trees in the grove today. After they are done, how many trees would be there?
Check: There would be 15 + 6 = 21 trees in total. This matches the original answer.

Q: If there are 3 cars in the parking lot and _____ more cars arrive, how many cars are in the parking lot?
A: 5
Rephrased: If there are 3 cars in the parking lot and x more cars arrive, there are 5 cars in the parking lot. Find the value of x.
Answer: There are originally 3 cars. x more cars arrive. 3 + x = 5, so x = 5 - 3 = 2. The answer is 2.
Final question: If there are 3 cars in the parking lot and 2 more cars arrive, how many cars are in the parking lot?
Check: There would be 3 + 2 = 5 cars in the parking lot. This matches the original answer.

...

Q: {{question}}
A: {{answer}}
Rephrased: 
\end{lstlisting}
\end{minipage}
\caption{Check your work with Rephrasing}
\label{prompt:rephrase_check_your_work}
\end{figure}

\begin{figure}[H]
\centering
\begin{minipage}{\columnwidth}
\begin{lstlisting}
You are given a math question with a blank value and an answer. Rephrase the given blanked question and answer pairs and then write a python function called solution() to find the value of x in the rephrased question. Return the value of x. You may assume the neccessary libraries are imported. Strictly follow the format given in the examples below, as the method will be executed with the same name.

Q: Ben has four boxes with ten basketball cards in each box. Ben received _____ cards from his mother. If he gives 58 cards to his classmates, how many cards does he has left?
A: 22
Rephrased: Ben has four boxes with ten basketball cards in each box. Ben received x cards from his mother. He gives 58 cards to his classmates. He has 22 cards left. Find the value of x.
Program:
```python
def solution():
    num_boxes = 4
    cards_per_box = 10
    total_cards_in_boxes = num_boxes * cards_per_box
    cards_from_mother = x
    cards_given_to_classmates = 58
    cards_left = 22
    
    equation = Eq(cards_from_mother + total_cards_in_boxes, cards_given_to_classmates + cards_left)
    blank = solve(equation)[0]

    return blank
```


Q: Natalia sold _____ clips to  her friends in April, and then she sold half as many clips in May. How many clips did Natalia sell altogether in April and May?
A: 72
Rephrased: Natalia sold x clips to her friends in April, and then she sold half as many clips in May. Natalia sells 72 clips altogether in April and May. Find the value of x.
Program:
```python
def solution():
    april_clips = x
    may_clips = april_clips / 2
    total_clips = 72
    
    equation = Eq(april_clips + may_clips, total_clips)
    blank = solve(equation)[0]

    return blank
```

Q: {{question}}
A: {{answer}}
Rephrased:
\end{lstlisting}
\end{minipage}
\caption{PAL-Tools with Rephrasing and Self-Refine: init prompt}
\label{prompt:self-refine-init}
\end{figure}

\begin{figure}[H]
\centering
\begin{minipage}{\columnwidth}
\begin{lstlisting}
You are given a question-answer pair with a blank, and a chain of thought (CoT) for filling in the blank. Go through the chain of thought step by step and point out mistakes, if any. Provide the final corrected answer as shown below.

Q: Kelly is grocery shopping at a supermarket and is making sure she has enough in her budget for the items in her cart. Her 5 packs of bacon cost $_____ in total and she has 6 packets of chicken which each cost twice as much as a pack of bacon. She also has 3 packs of strawberries, priced at $4 each, and 7 packs of apples, each priced at half the price of a pack of strawberries. If Kelly's budget is $65 then how much money, in dollars, does she have left in her budget?
A: 5
Rephrased: Kelly is grocery shopping at a supermarket and is making sure she has enough in her budget for the items in her cart. Her 5 packs of bacon cost a total of x dollars, and she has 6 packets of chicken, each costing twice as much as a pack of bacon. She also has 3 packs of strawberries priced at $4 each, and 7 packs of apples, each priced at half the price of a pack of strawberries. If Kelly's budget is $65, then how much money, in dollars, does she have left in her budget?

Program:

```python
def solution():
    bacon_cost = x
    chicken_cost = 2 * bacon_cost
    strawberry_cost = 4
    apple_cost = strawberry_cost / 2
    budget = 65
    remaining_budget = budget - (5 * bacon_cost + 6 * chicken_cost + 3 * strawberry_cost + 7 * apple_cost)
    
    equation = Eq(remaining_budget, 5)
    blank = solve(equation)[0]

    return blank
```

Correction:

The rephrased problem does not end with 'Find the value of x'. Instead, it asks us to find the money Kelly has left in her budget. This is incorrect.

Let's check the program:

    bacon_cost = x
    chicken_cost = 2 * bacon_cost

Looks good

    strawberry_cost = 4
    apple_cost = strawberry_cost / 2

Looks good

    budget = 65
    remaining_budget = budget - (5 * bacon_cost + 6 * chicken_cost + 3 * strawberry_cost + 7 * apple_cost)

The equation formed is correct, as are the number of packs of bacon, chicken, strawberry and apples

    equation = Eq(remaining_budget, 5)
    blank = solve(equation)[0]
    return blank

The equation has been equated to the correct answer, and the program returns the blank.

The program provided is completely correct!

\end{lstlisting}
\end{minipage}
\caption{PAL-Tools with Rephrasing and Self-Refine: feedback prompt}
\label{prompt:self-refine-feedback}
\end{figure}

\begin{figure}[H]
\centering
\begin{minipage}{\columnwidth}
\begin{lstlisting}
Final Rephrased Problem:
Kelly is grocery shopping at a supermarket and is making sure she has enough in her budget for the items in her cart. Her 5 packs of bacon cost a total of x dollars, and she has 6 packets of chicken, each costing twice as much as a pack of bacon. She also has 3 packs of strawberries priced at $4 each, and 7 packs of apples, each priced at half the price of a pack of strawberries. If Kelly's budget is $65, then she has $5 left in her budget after shopping. Find the value of x.

Final Program:
```python
def solution():
    bacon_cost = x
    chicken_cost = 2 * bacon_cost
    strawberry_cost = 4
    apple_cost = strawberry_cost / 2
    budget = 65
    remaining_budget = budget - (5 * bacon_cost + 6 * chicken_cost + 3 * strawberry_cost + 7 * apple_cost)
    
    equation = Eq(remaining_budget, 5)
    blank = solve(equation)[0]

    return blank
```
...

Q: {{question}}
A: {{answer}}
Rephrased: {{rephrased}}
Program:
```python
{{program}}
```

Correction:  

\end{lstlisting}
\end{minipage}
\caption{PAL-Tools with Rephrasing and Self-Refine: feedback prompt continued}
\label{prom:self-refine-feedback_c}
\end{figure}

\begin{figure}[H]
\centering
\begin{minipage}{\columnwidth}
\begin{lstlisting}
Solve the given question step by step and check whether the answer obtained matches the given answer. Follow the format specified in the examples below:

Q: There are 15 trees in the grove. Grove workers will plant 6 trees in the grove today. After they are done, how many trees would be there?
A: 21 
Check: There would be 15 + 6 = 21 trees in total. The given answer is 21. This matches the given answer.

Q: If there are 3 cars in the parking lot and 2 more cars arrive, how many cars are in the parking lot?
A: 5
Check: There would be 3 + 2 = 5 cars in the parking lot. The answer is 5. This matches the given answer.

...

Q: {{question}}
A: {{answer}}
Check: 

\end{lstlisting}
\end{minipage}
\caption{Verifier prompt for ensembling}
\label{prompt:verifier}
\end{figure}

\begin{figure}[H]
\centering
\begin{minipage}{\columnwidth}
\begin{lstlisting}
Find a phrase to fill in the blank given the question and answer. Your answer should contain a decimal. Follow the format specified in the examples below and solve step by step:

Q: There are 15 trees in the grove. _____. After they are done, how many trees would be there?
A: 21 
Answer: There are 15 trees originally, Then there are going to be 21 trees in the end. So there must have been 21 - 15 = 6 trees added.
Phrase: Grove workers will plant 6 trees today

Q: If there are 3 cars in the parking lot and _____, how many cars are in the parking lot?
A: 5
Answer: There are originally 3 cars. There are finally 5 cars, so 5 - 3 = 2 cars arrived.
Phrase: 2 cars arrive

...

Q: {{question}}
A: {{answer}}
Answer:
\end{lstlisting}
\end{minipage}
\caption{CoT for phrase-masked task}
\label{prompt:cot_phrase_masked}
\end{figure}

\begin{figure}[H]
\centering
\begin{minipage}{\columnwidth}
\begin{lstlisting}
Given blanked question Q, we need to complete it such that its answer is A.
First guess the blank. Then rephrase the question to find x by putting value in A. Then solve it step-by-step. Follow the format specified in the examples below:

Q: There are 15 trees in the grove. _____. After they are done, how many trees would be there?
A: 21
Guess: x trees were added.
Rephrased: There are 15 trees in the grove. x trees were added. After they are done, there would be 21 trees. Find the value of x.
Answer: There are 15 trees originally, Then there were 21 trees after some more were planted. So there must have been x = 21 - 15 = 6 trees. The answer is 6.

Q: If there are 3 cars in the parking lot and _____, how many cars are in the parking lot?
A: 5
Guess: x more cars arrive
Rephrased: If there are 3 cars in the parking lot and x more cars arrive, how many cars are in the parking lot?
Answer: There are originally 3 cars. x more cars arrive. 3 + x = 5, so x = 5 - 3 = 2. The answer is 2.

...

Q: {{question}}
A: {{answer}}
\end{lstlisting}
\end{minipage}
\caption{CoT Rephrase for phrase-masked task}
\label{prompt:cot_rephrase_phrase_masked}
\end{figure}

\begin{figure}[H]
\centering
\begin{minipage}{\columnwidth}
\begin{lstlisting}
Given blanked question Q, we need to complete it such that its answer is A.
First guess the blank by making use of unknown x. Then solve it to find x step-by-step. Check your work by substituting your answer in the blank, solving the question and comparing to the original answer. Follow the format specified in the examples below:

Q: There are 15 trees in the grove. _____. After they are done, how many trees would be there?
A: 21
Guess: x trees were added. 
Answer: There are 15 trees originally, Then there were 21 trees after some more were planted. So there must have been x = 21 - 15 = 6 trees. The answer is 6.
Final question: There are 15 trees in the grove. 6 trees were added. After they are done, how many trees would be there?
Check: There would be 15 + 6 = 21 trees in total. The original answer was 21. This matches the original answer.

Q: If there are 3 cars in the parking lot and _____, how many cars are in the parking lot?
A: 5
Guess: x more cars arrive
Answer: There are originally 3 cars. x more cars arrive. 3 + x = 5, so x = 5 - 3 = 2. The answer is 2.
Final question: If there are 3 cars in the parking lot and 2 more cars arrive, how many cars are in the parking lot?
Check: There would be 3 + 2 = 5 cars in the parking lot. The original answer was 5. This matches the original answer.

...

Q: {{question}}
A: {{answer}}
Guess:
\end{lstlisting}
\end{minipage}
\caption{Check-your-work for phrase-masked task}
\label{prompt:cyw_phrase_masked}
\end{figure}

\begin{figure}[H]
\centering
\begin{minipage}{\columnwidth}
\begin{lstlisting}
Given blanked question Q, we need to complete it such that its answer is A.
First, guess the blank by making use of unknown x. Rephrase the question to find x by making use of the value of A. Then solve it step b step to find the value of blank. Strictly follow the format given in the examples below.

Q: Ben has four boxes with ten basketball cards in each box. _____. If he gives 58 cards to his classmates, how many cards does he have left?
A: 22

Guess: He gets x more cards
Rephrased: Ben has four boxes with ten basketball cards in each box. He gets x more cards. He gives 58 cards to his classmates. He has 22 cards left. Find the value of x.
Peano solution:


Let a be number of boxes [[var a]]. We have [[eq a = 4]].
Let b be number of cards in each box [[var b]]. We have [[eq b = 10]].
Let c be number of cards Ben initially has [[var c]]. We have [[eq c = a * b]].
Let x be cards he got [[var x]].
Let d be total cards with Ben [[var d]]. We have [[eq d = c + x]]
Let e be cards given to classmates [[var e]]. We have [[eq e = 58]].
Let f be cards left [[var f]]. From given Answer, we have [[eq f = 22]]. 
As cards left are also total cards - cards given to classmates, we have [[eq f = d - e]]
The answer will be the value of x [[answer x]].





Q: Natalia sold 48 clips to her friends in April, _____. How many clips did Natalia sell altogether in April and May?
A: 72

Guess: and sold x in May
Rephrased: Natalia sold 48 clips to her friends in April, and sold x in May. Natalia sells 72 clips altogether in April and May. Find the value of x.
Peano solution:


Let x be number of clips Natalia sold in April [[var a]]. We have [[eq a = 48]].
Let a be number of clips Natalia sold in May [[var x]].
Let b be number of clips sold altogether [[var b]]. From given Answer, we have [[eq b = 72]].
As clips sold altogether are also the sum of clips sold in April and May, we have [[eq b = x + a ]]
The answer will be the value of x [[answer x]].

...

Q: {{question}}
A: {{answer}}
Guess:
\end{lstlisting}
\end{minipage}
\caption{Rephrase Tools for phrase-masked task}
\label{prompt:r_tools_phrase_masked}
\end{figure}

\begin{figure}[H]
\centering
\begin{minipage}{\columnwidth}
\begin{lstlisting}
Given blanked question Q, we need to complete it such that its answer is A.
First, guess the blank by making use of unknown x. Rephrase the question to find x by making use of the value of A. Then write a Python function called solution() using sympy that assumes the value of the blank is x and creates an equation in x that is solved by sympy.solve. Return the value of the blank. You may assume the necessary libraries are imported. Strictly follow the format given in the examples below, as the method will be executed with the same name.

Q: Ben has four boxes with ten basketball cards in each box. _____. If he gives 58 cards to his classmates, how many cards does he have left?
A: 22
Guess: He gets x more cards
Rephrased: Ben has four boxes with ten basketball cards in each box. He gets x more cards. He gives 58 cards to his classmates. He has 22 cards left. Find the value of x.
Program:
```python
def solution():
    num_boxes = 4
    cards_per_box = 10
    total_cards_in_boxes = num_boxes * cards_per_box
    cards_got = x
    cards_given_to_classmates = 58
    cards_left = 22
    
    equation = Eq(cards_got + total_cards_in_boxes, cards_given_to_classmates + cards_left)
    blank = solve(equation)[0]

    return blank
```

Q: Natalia sold 48 clips to her friends in April, _____. How many clips did Natalia sell altogether in April and May?
A: 72
Guess: and sold x in May
Rephrased: Natalia sold 48 clips to her friends in April, and sold x in May. Natalia sells 72 clips altogether in April and May. Find the value of x.
Program:
```python
def solution():
    april_clips = 48
    may_clips = x
    total_clips = 72
    
    equation = Eq(april_clips + may_clips, total_clips)
    blank = solve(equation)[0]

    return blank
```
...

Q: {{question}}
A: {{answer}}
Guess:
\end{lstlisting}
\end{minipage}
\caption{Rephrase Pal-Tools for phrase-masked task}
\label{prompt:r_pal_tools_phrase_masked}
\end{figure}

\end{document}